# Comparison of different convolutional neural network activation functions and methods for building ensembles

Loris Nanni [1],[*] Gianluca Maguolo[1], Sheryl Brahnam[2], and Michelangelo Paci[3]

[1] DEI, University of Padua, Via Gradenigo 6, 35131 Padova, Italy; loris.nanni@unipd.it and gianluca.maguolo@phd.unipd.it
[2] Department of Information Technology and Cybersecurity, Missouri State University, 901 S. National Street, Springfield, MO 65804, USA; sbrahnam@missouristate.edu
[3] BioMediTech, Faculty of Medicine and Health Technology, Tampere University, Arvo Ylpön katu 34, D 219, FI-33520, Tampere, Finland; michelangelo.paci@tuni.fi
[*] Correspondence: loris.nanni@unipd.it

**Abstract:** Recently, much attention has been devoted to finding highly efficient and powerful activation functions for CNN layers. Because activation functions inject different nonlinearities between layers that affect performance, varying them is one method for building robust ensembles of CNNs. The objective of this study is to examine the performance of CNN ensembles made with different activation functions, including six new ones presented here: 2D Mexican ReLU, TanELU, MeLU+GaLU, Symmetric MeLU, Symmetric GaLU, and Flexible MeLU. The highest performing ensemble was built with CNNs having different activation layers that randomly replaced the standard ReLU. A comprehensive evaluation of the proposed approach was conducted across fifteen biomedical data sets representing various classification tasks. The proposed method was tested on two basic CNN architectures: Vgg16 and ResNet50. Results demonstrate the superiority in performance of this approach. The MATLAB source code for this study will be available at https://github.com/LorisNanni.

## 1. Introduction

First developed in the 1940s, artificial neural networks have had a checkered history, sometimes lauded by researchers for their unique computational powers and other times discounted for being no better than statistical methods. About a decade ago, artificial neural networks called deep learners composed of multiple specialized hidden layers radically changed the direction of machine learning and rapidly made significant inroads into many scientific and engineering areas [1-5]. The strength of deep learners is illustrated by the many successes achieved by one of the most famous and robust deep learning architectures, Convolutional Neural Networks (CNNs). CNNs frequently win image recognition competitions and have consistently outperformed other classifiers on a range of image classification tasks [1, 6]: object detection [7], face recognition [8], and machine translation [9], to name but a few. Not only do CNNs continue to eclipse traditional classifiers, but they have also been shown to outperform human beings, including experts, in many image recognition tasks. CNNs outshine humans beings, for example, in recognizing faces [10, 11], traffic signs [12], handwritten digits [12, 13], and the fourteen million objects categorized into one thousand classes in the ImageNet data set [14, 15].

Evolutions in CNN design initially centered around building better network topologies. Because activation functions impact training dynamics and performance, many researchers have also focused on developing better activation functions. For many years, the sigmoid and the hyperbolic tangent were the most popular neural network activation functions. The hyperbolic tangent's main advantage over the sigmoid is that the hyperbolic has a steeper derivative than the sigmoid function. Neither function, however, works that well with deep learners since both are subject to the vanishing gradient problem. It was soon realized that nonlinearities function better with deep learners. One of the first nonlinearities to demonstrate improved performance with CNNs was the now-classic activation function Rectified Linear Units (ReLU) [16], which is equal to the identity function with positive input and zero with negative input [17]. Although ReLU is non-differentiable, it gave AlexNet the edge to win the 2012 ImageNet competition [1].

The success of ReLU in AlexNet motivated researchers to investigate other nonlinearities and the desirable properties they possess. As a consequence, variations of ReLU have proliferated. For example, Leaky ReLU [18], like ReLU, is also equivalent to the identity function for positive values but has a hyperparameter $\alpha > 0$ applied to the negative inputs to ensure the gradient is never zero. As a result, Leaky ReLU is not as prone to getting caught in local minima and counters ReLU's problem with hard zeros that makes it more likely to fail to activate. Similar to Leaky ReLU is Exponential Linear Units (ELU) [19]. The advantage offered by ELU derives from the fact that it always produces a positive gradient since it exponentially decreases to the limit point $\alpha$ as the input goes to minus infinity. A disadvantage of ELU, however, is that it saturates on the left side. Another activation function designed to handle the vanishing gradient problem is the Scaled Exponential Linear Unit (SELU) [20]. SELU is identical to ELU except that it is multiplied by the constant $\lambda > 1$ to maintain the mean and the variance of the input features.

Until 2015, activation functions were engineered to modify the weights and biases of a neural network. Parametric ReLU (PReLU) [14] gave Leaky ReLU a learnable parameter applied to the negative slope. The success of PReLU set in motion more research into learnable activation functions [21, 22]. The Adaptive Piecewise Linear Unit (APLU) [21], for instance, independently learns during the training phase the piecewise slopes and points of nondifferentiability for each neuron using gradient descent. In consequence, it can imitate any piecewise linear function.

Aside from applying a learnable parameter to part of an activation function, as with PReLu and APLU, the construction of an activation function from a fixed set of them can be learned as well. In [23], for instance, an activation function was produced that automatically learned the best combinations of tanh, ReLU, and the identity function. Another activation function of this type is the S-shaped Rectified Linear activation Unit (SReLU) [24]. Using reinforcement learning, SReLU was designed to learn convex and non-convex functions to imitate both the Webner-Fechner and the Stevens law. This process produced an activation called Swish, which the authors view as a smooth function that nonlinearly interpolates between the linear function and ReLU.

Similar to APLU is the Mexican ReLU (MeLU [25]), whose shape resembles the Mexican hat wavelet. MeLU is a piecewise linear activation function that combines PReLU with many Mexican hat functions. Like APLU, MeLU has learnable parameters that approximate the same piecewise linear functions equivalent to identity when $x$ is sufficiently large, but MeLU differs from APLU, first, in having a much larger number of parameters (collectively called a hyperparameter), second, in the manner in which the approximations are calculated for each function, and, third, in the gradients.

Combining different activation functions has recently been shown to be a highly effective way to train robust classifier systems. In [26], CNNs with different activation functions were trained and fused; and, in [27], different activation functions were inserted into the layers of a single network. Both methods produced excellent results and surpassed the performance of the single CNNs.

In this paper, we extend [26] by comparing a large set of seventeen activation functions using two CNNs, Vgg16 [28], and ResNet50 [29], across fifteen biomedical data sets representing different biomedical tasks. The set of activation functions include the state-of-the-art and six new ones (2D Mexican ReLU, TanELU, MeLU+GaLU, Symmetric MeLU, Symmetric GaLU, and Flexible MeLU) proposed here. Also compared here are different methods for generating the CNN ensembles. The best performance is obtained by randomly replacing every ReLu layer of each CNN with a different activation function.

This paper is organized as follows. In Section 2, all the tested activation functions, including the new ones presented here, are described. In section 3, the stochastic approach for constructing CNN ensembles is detailed (the other methods are described in the experimental section). In Section 4, the evaluation of the activation functions on two CNN architectures is reported and discussed. Finally, in Section 5, the paper is concluded with ideas for future investigations.

## 2. Activation functions used with the two CNNs

Some of the best performing activation functions were selected as candidates to be substituted into two of the most highly regarded CNN architectures: Vgg16 [28] and ResNet50 [29], each pre-trained on ImageNet. VGG16 [28], also known as the OxfordNet, is the second-place winner in the ILSVRC 2014 competition and was one of the deepest neural networks produced at that time. The input into VGG16 passes through stacks of convolutional layers, with filters having small receptive fields. Stacking these layers is similar in effect to CNNs having larger convolutional filters, but the stacks involve fewer parameters and are thus more efficient. ResNet50 [6], winner of ILSVRC 2015 contest and now a popular network, is a CNN with fifty layers known for its skip connections that sum the input of a block to its output, a technique that promotes gradient propagation and that preserves lower semantic information so that higher layers can work on it.

The remainder of this section mathematically describes and discusses when needed the twenty activation functions investigated in this study: ReLU [16], Leaky ReLU [30], ELU [19], SELU [20], PReLU [14], APLU [21], SReLU [31], MeLU [25], Splash [32], Mish [33], PDELU [34], Swish [24], Soft Learnable [35], SRS [35] and GaLU ([36]), as well as the novel activation functions proposed here: 2D Mexican ReLU, TanELU, MeLU + GaLU, Symmetric MeLU, Symmetric GaLU and Flexible MeLU.

### 2.1. ReLu

ReLU [16], illustrated in Figure 1, is defined as

$$y_i = f(x_i) = \begin{cases} 0, & x_i < 0 \\ x_i, & x_i \geq 0. \end{cases}$$

The gradient of ReLU is

$$\frac{dy_i}{dx_i} = f'(x_i) = \begin{cases} 0, & x_i < 0 \\ 1, & x_i \geq 0. \end{cases}$$

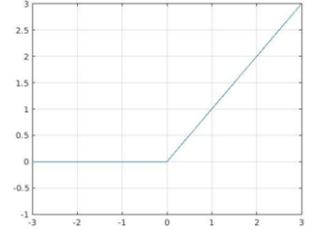

**Figure 1. ReLU**

### 2.2. Leaky ReLU

In contrast to ReLu, Leaky ReLu [18] has no point with a null gradient. Leaky ReLU, illustrated in Figure 2, is defined as

$$y_i = f(x_i) = \begin{cases} ax_i, & x_i < 0 \\ x_i, & x_i \geq 0, \end{cases}$$

where $a$ (set to 0.01 here) is a small real number.

The gradient of Leaky ReLU is

$$\frac{dy_i}{dx_i} = f'(x_i) = \begin{cases} a, & x_i < 0 \\ 1, & x_i \geq 0. \end{cases}$$

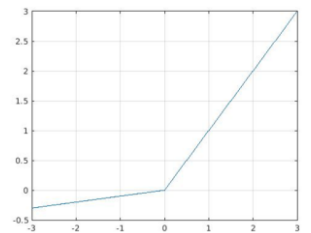

**Figure 2. Leaky ReLU**

## 2.3. ELU

Exponential Linear Unit (ELU) [19] is differentiable; and, as is the case with Leaky ReLU, the gradient is always positive and bounded from below by $-a$. ELU, illustrated in Figure 3, is defined as

$$y_i = f(x_i) = \begin{cases} a(\exp(x_i) - 1), & x_i < 0 \\ x_i, & x_i \geq 0, \end{cases}$$

where $a$ (set to 1 here) is a real number.

The gradient of Leaky ELU is

$$\frac{dy_i}{dx_i} = f'(x_i) = \begin{cases} a \exp(x_i), & x_i < 0 \\ 1, & x_i \geq 0. \end{cases}$$

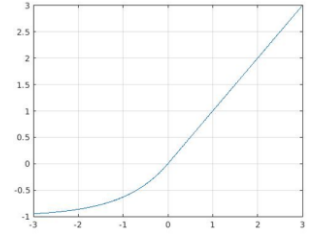

**Figure 3. ELU**

## 2.4. PReLU

Parametric ReLU (PReLU) [14] is identical to Leaky ReLU except that the parameter $a_c$ (different for every channel of the input) is learnable. PReLU is defined as

$$y_i = f(x_i) = \begin{cases} a_c x_i, & x_i < 0 \\ x_i, & x_i \geq 0, \end{cases}$$

where $a_c$ is a real number.

The gradients of PReLU are

$$\frac{dy_i}{dx_i} = f'(x_i) = \begin{cases} a_c, & x_i < 0 \\ 1, & x_i \geq 0, \end{cases}$$

$$\frac{dy_i}{da_c} = \begin{cases} x_i, & x_i < 0 \\ 0, & x_i \geq 0. \end{cases}$$

Slopes on the left-hand sides are all initialized to 0.

## 2.5. SReLU

S-Shaped ReLU (SReLU) [31] is composed of three piecewise linear functions expressed by four learnable parameters ($t^l, t^r, a^l$, and $a^r$ initialized as $a^l = 0, t^l = 0$, $t^r = maxInput$, a hyperparameter). This rather large set of parameters gives SReLU its high representational power. SReLU, illustrated in Figure 4, is defined as

$$y_i = f(x_i) = \begin{cases} t^l + a^l(x_i - t^l), & x_i \leq t^l \\ x_i, & t^l < x_i < t^r, \\ t^r + a^r(x_i - t^r), & x_i \geq t^r. \end{cases}$$

where $a_c$ is a real number.

The gradients of SeLU are

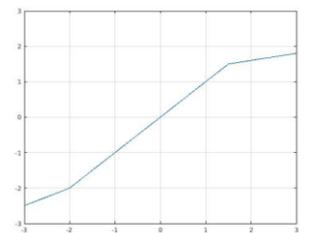

**Figure 4. SReLU.**

$$\frac{dy_i}{dx_i} = f'(x_i) = \begin{cases} a^l, & x_i \leq t^l \\ 1, & t^l < x_i < t^r, \\ a^r, & x_i \geq t^r \end{cases}$$

$$\frac{dy_i}{da^l} = \begin{cases} x_i - t^l, & x_i \leq t^l \\ 0, & x_i > t^l, \end{cases}$$

$$\frac{dy_i}{dt^l} = \begin{cases} 1 - a^l, & x_i \leq t^l \\ 0, & x_i > t^l, \end{cases}$$

$$\frac{dy_i}{da^r} = \begin{cases} x_i - t^r, & x_i \geq t^r \\ 0, & x_i < t^r, \end{cases}$$

$$\frac{dy_i}{dt^r} = \begin{cases} 1 - a^r, & x_i \geq t^r \\ 0, & x_i < t^r. \end{cases}$$

Here we use $a^l = 0.5$, $a^r = 0.2$, $t^l = -2$, $t^r = 1.5$.

### 2.6. APLU

Adaptive Piecewise Linear Unit (APLU) [21] is a linear piecewise function that can approximate any continuous function on a compact set. The gradient of APLU is the sum of the gradients of ReLU and of the functions contained in the sum. APLU is defined as

$$y_i = \text{ReLU}(x_i) + \sum_{c=1}^{n} a_c \min(0, -x_i + b_c),$$

where $a_c$ and $b_c$ are real numbers that are different for each channel of the input.

With respect to the parameters $a_c$ and $b_c$, the gradients of APLU are

$$\frac{df(x,a)}{da_c} = \begin{cases} -x + b_c, & x < b_c \\ 0, & x \geq b_c \end{cases}$$

$$\frac{df(x,a)}{db_c} = \begin{cases} a_c, & x < b_c \\ 0, & x \geq b_c. \end{cases}$$

The values for $a_c$ are initialized here to zero, with points randomly initialized. The 0.001 $L^2$-penalty is added to the norm of the $a_c$ values. This addition requires that another term $L^{reg}$ be included in the loss function:

$$L^{reg} = \sum_{c=1}^{n} |a_c|^2.$$

Furthermore, a relative learning rate is added: $maxInput$ multiplied by the smallest value used for the rest of the network. If $\lambda$ is the global learning rate, then the learning rate $\lambda^*$ of the parameters $a_c$ would be

$$\lambda^* = \lambda/maxInput.$$

### 2.7. MeLU

The mathematical basis of the Mexican ReLU (MeLu) [25] activation function can be described as follows. Given the real numbers $a$ and $\lambda$ and letting $\phi^{a,\lambda}(x) = \max(\lambda - |x - a|, 0)$ be a so-called *Mexican hat* type of function, then when $|x - a| > \lambda$, the function $\phi^{a,\lambda}(x)$ is null but increases with a derivative of 1 and $a$ between $a - \lambda$ and decreases with a derivative of $-1$ between $a$ and $a + \lambda$.

Considering the above, MeLU is defined as

$$y_i = MeLU(x_i) = PReLU^{c_0}(x_i) + \sum_{j=1}^{k-1} c_j \phi^{a_j,\lambda_j}(x_i),$$

where $k$ is the number of learnable parameters for each channel, $c_j$ are the learnable parameters, and $c_0$ is the vector of parameters in PReLU.

The parameter $k$ ($k = 4,8$ here) has one value for PReLU and $k - 1$ values for the coefficients in the sum of the Mexican hat functions. The real numbers $a_j$ and $\lambda_j$ are fixed (see Table 1) and are chosen recursively. The value of $maxInput$ is set to 256. The first

Mexican hat function has its maximum at $2 \cdot maxInput$ and is equal to zero in 0 and $4 \cdot maxInput$. The next two functions are chosen to be zero outside the interval $[0, 2 \cdot maxInput]$ and $[2 \cdot maxInput, 4 \cdot maxInput]$, with the requirement being they have their maximum in $maxInput$ and $3 \cdot maxInput$.

**Table 1**. Fixed parameters of MeLU with $maxInput = 256$ (These are the same values as in [25]).

| j | 1 | 2 | 3 | 4 | 5 | 6 | 7 |
|---|---|---|---|---|---|---|---|
| $a_j$ | 512 | 256 | 768 | 128 | 384 | 640 | 896 |
| $\lambda_j$ | 512 | 256 | 256 | 128 | 128 | 128 | 128 |

The Mexican hat functions that MeLU is based on are continuous and piecewise differentiable. Mexican hat functions are also a Hilbert basis on a compact set with the $L^2$ norm. As a result, MeLU can approximate every function in $L^2([0,1024])$ as $k$ goes to infinity.

When the $c_i$ learnable parameters are set to zero, MeLU is identical to ReLU. Thus, MeLU can easily replace networks pre-trained with ReLU. This is not to say, of course, that MeLU cannot replace the activation functions of networks trained with Leaky ReLU and PReLU. In this study, all $c_i$ are initialized to zero, so start off as ReLU, with all its attendant properties.

MeLU's hyperparameter ranges from zero to infinity, producing many desirable properties. The gradient is rarely flat, and in no direction does saturation take place. As the size of the hyperparameter approaches infinity, it can approximate every continuous function on a compact set. Finally, the modification of any given parameter only changes the activation on a small interval and only when needed, making optimization relatively simple.

*2.8. GaLU*

Piecewise linear odd functions, composed of many linear pieces, do a better job of approximating nonlinear functions than does ReLU [37]. For this reason, Gaussian ReLU (GaLU) [36], based on Gaussian types of functions, aims to add more linear pieces than does MeLU. Since GaLU extends MeLU, GaLU retains all the favorable properties discussed in section 2.7.

Letting $\phi_g^{a,\lambda}(x) = \max(\lambda - |x - a|, 0) + \min(|x - a - 2\lambda| - \lambda, 0)$ be a Gaussian type of function, where $a$ and $\lambda$ are real numbers, GaLU is defined, similarly to MeLU, as

$$y_i = GaLU(x_i) = PReLU^{c_0}(x_i) + \sum_{j=1}^{k-1} c_j\, \phi_g^{a_j,\lambda_j}(x_i).$$

In this work, $k = 2$ parameters for what will be called in the experimental section *Small GaLU* and $k = 4$ for GaLU proper.

Like MeLU, GaLU has the same set of fixed parameters. A comparison of values for the fixed parameters with $maxInput = 1$ is provided in Table 2.

**Table 2**. Comparison of the Fixed parameters of GaLU and MeLU with $maxInput = 1$.

| | | j | 1 | 2 | 3 | 4 | 5 | 6 | 7 |
|---|---|---|---|---|---|---|---|---|---|
| MeLU | $a_j$ | | 2 | 1 | 3 | 0.5 | 1.5 | 2.5 | 3.5 |
| | $\lambda_j$ | | 2 | 1 | 1 | 0.5 | 0.5 | 0.5 | 0.5 |
| GaLU | $a_j$ | | 1 | 0.5 | 2.5 | 0.25 | 1.25 | 2.25 | 3.25 |
| | $\lambda_j$ | | 1 | 0.5 | 0.5 | 0.25 | 0.25 | 0.25 | 0.25 |

*2.9. PDELU*

Piecewise linear Parametric Deformable Exponential Linear Unit (PDELU) [34] is designed to have zero mean, which speeds up the training process. It is defined as

$$y_i = f(x_i) = \begin{cases} x_i, & x_i > 0 \\ a_i \cdot \left([1 + (1-t)x_i]_+^{\frac{1}{1-t}} - 1\right), & x_i \leq 0. \end{cases}$$

### 2.10. Swish

Swish [24] is designed using reinforcement learning to learn to efficiently sum, multiply, and compose different functions that are used as building blocks. The best function is

$$y = f(x) = x \cdot sigmoid(\beta x) = \frac{x}{1+e^{-\beta x}},$$

where $\beta$ (set here to 1) is a parameter that can be learnable during training.

### 2.11. Mish

Mish [33] is defined as

$$y = f(x) = x \cdot tanh(softplus(\alpha x)) = x \cdot tanh(ln(1 + e^{\alpha x})),$$

where $\alpha$ is a learnable parameter.

### 2.12. SRS

Soft Root Sign (SRS) [35] is defined as

$$y = f(x) = \frac{x}{\frac{x}{\alpha} + e^{-\frac{x}{\beta}}},$$

where $\alpha$ and $\beta$ are non-negative learnable parameters. The output has zero means if the input is a standard normal.

### 2.13. Soft Learnable

Soft Learnable [35] is a very recent activation function defined as

$$y = f(x) = \begin{cases} x, & x > 0 \\ \alpha \cdot ln\left(\frac{1+e^{\beta x}}{2}\right), & x \leq 0, \end{cases}$$

where $\alpha$, $\beta$ are positive parameters. We used two different versions of this activation, depending on whether the parameter $\beta$ is fixed (labeled here as SoftLearnable) or not (labeled here as SoftLearnable2).

### 2.14. Splash

Splash [32] is another modification of APLU that makes the function symmetric. In the definition of APLU, let $a_i$ and $b_i$ be the learnable parameters $APLU_{a_i,b_i}(x)$. Then Splash is defined as

$$Splash_{a_i^+,a_i^-,b_i}(x) = APLU_{a_i^+,b_i}(x) + APLU_{a_i^-,b_i}(-x).$$

This equation's hinges are symmetric with respect to the origin. The authors in [32] claim that this network is more robust against adversarial attacks.

### 2.15. 2D MeLU (New)

2D Mexican ReLU (2D MeLU) is an activation function presented here that is not defined component-wise; instead, every output neuron depends on two input neurons. If a layer has $N$ neurons (or channels), its output is defined as

$$y_i = PReLU^{c_0}(x_i) + PReLU^{c_0}(x_{i+1}) + \sum_{u,v=1}^{k-1} c_j \, \phi^{a_{u,v}, \lambda_{\max(u,v)}}(x_i, x_{i+1}),$$

where $\phi^{a_j,\lambda_j}(x_i, x_{i+1}) = \max(\lambda_j - |(x_i, x_{i+1}) - a_{u,v}|, 0)$.

The parameter $a_{u,v}$ is a two-dimensional vector whose entries are the same as those used in MeLU. In other words, $a_{u,v} = (a_u, a_v)$ as defined in Table 1. Likewise, $\lambda_{\max(u,v)}$ is defined as it is for MeLU in Table 1.

*2.16. TanELU (New)*

TanELU is an activation function presented here that is simply the weighted sum of tanh and ReLU:

$$y_i = ReLU(x_i) + a_i \tanh(x_i),$$

where $a_i$ is a learnable parameter.

*2.17. MeLU + GaLU (New)*

MeLU + GaLU is an activation function presented here that is, as its name suggests, the weighted sum of MeLU and GaLU:

$$y_i = (1 - a_i)MeLU(x_i) + a_i\, GaLU(x_i),$$

where $a_i$ is a learnable parameter.

*2.18. Symmetric MeLU (New)*

Symmetric MeLU is the equivalent of MeLU, but it is symmetric like Splash. Symmetric MeLU is defined as

$$y_i = MeLU(x_i) + MeLU(-x_i),$$

where the coefficients of the two MeLUs are the same. In other words, the $k$ coefficients of $MeLU(x_i)$ are the same as $MeLU(-x_i)$.

*2.19. Symmetric GaLU (New)*

Symmetric GaLU is the equivalent of symmetric MeLU but uses GaLU instead of MeLU. Symmetric GaLU is defined as

$$y_i = GaLU(x_i) + GaLU(-x_i),$$

where the coefficients of the two GaLUs are the same. In other words, the $k$ coefficients of $GaLU(x_i)$ are the same as $GaLU(-x_i)$.

*2.20. Flexible MeLU (New)*

Flexible MeLU is a modification of MeLU where the peaks of the Mexican function are also learnable. This feature makes it more similar to APLU since its points of non-differentiability are also learnable. With respect to MeLU, APLU has more hyperparameters.

**3. Methods for combining CNNs**

One of the objectives of this study is to use several methods for combining the two CNNs with the different activation functions. Two methods are in need of discussion: Sequential Forward Floating Selection (SFFS) [38] and the stochastic method for combining CNNs introduced in [27].

*3.1. Sequential Forward Floating Selection (SFFS)*

A popular method for selecting an optimal set of descriptors, SFFS [38], has been adapted for selecting the most performing/independent classifiers to be added to the ensemble. In applying the SFFS method, each model to be included in the final ensemble is selected by adding, at each step, the model which provides the highest increment in performance compared to the existing subset of models. Then a backtracking step is performed to exclude the worst model from the actual ensemble.

This method for combining CNNs is labeled *Selection* in the experimental section. Since SFFS requires a training phase, we perform a leave-one-out-data set selection to select the best-suited models.

*3.2. Stochastic method*

The stochastic approach [27] involves randomly substituting all the activations in a CNN architecture with a new one selected from a pool of potential candidates. Random selection is repeated many times to generate a set of networks that will be fused together. The candidate activation functions within a pool differ depending on the CNN architecture. Some activation functions appear to perform poorly and some quite well on a given CNN, with quite a large variance. The activation functions included in the pools for each of the CNNs tested here are provided in Table 3. The CNN ensembles randomly built from these pools varied in size, as will be noted in the experimental section, which investigates the different ensembles. Ensemble decisions are combined by sum rule, where the softmax probabilities of a sample given by all the networks are averaged, and the new score is used for classification.

The stochastic method of combining CNNs is labeled Stoc in the experimental section.

**Table 3**. Activation functions included (✓) in the pools for each of the two CNN architectures.

| Activation | VGG16 | ResNet50 |
|---|---|---|
| MeLU (k=8) | ✓ | ✓ |
| Leaky ReLu | ✓ | ✓ |
| ELU | ✓ | ✓ |
| MeLU (k=4) | ✓ | ✓ |
| PReLU | ✓ | ✓ |
| SReLU | ✓ | ✓ |
| APLU | ✓ | ✓ |
| ReLu | ✓ | ✓ |
| Small GaLU | ✓ | ✓ |
| GaLU | ✓ | ✓ |
| Flexible MeLU | ✓ | ✓ |
| TanELU | ✓ | ✓ |
| 2D MeLU |  | ✓ |
| MeluGalu | ✓ | ✓ |
| Splash |  | ✓ |
| Symmetric Galu | ✓ | ✓ |
| Symmetric Melu | ✓ | ✓ |
| Soft Learnable v2 | ✓ | ✓ |
| Soft Learnable | ✓ | ✓ |
| PDELU | ✓ | ✓ |
| Mish | ✓ | ✓ |
| SRS |  | ✓ |
| Swish Learnable | ✓ | ✓ |
| Swish | ✓ | ✓ |

## 4. Experimental Results

*4.1. Biomedical data sets*

Each of the activation functions detailed in section 2 is tested on the CNNs using the following fifteen publicly available biomedical data sets:

1. CH (CHO data set [39]): this is a data set containing 327 fluorescence microscope images of Chinese Hamster Ovary cells divided into five classes: an-ti-giantin, Hoechst 33258 (DNA), anti-lamp2, anti-nop4, and anti-tubulin.
2. HE (2D HeLa data set [39]): this is a balanced data set containing 862 fluorescence microscopy images of HeLa cells stained with various organelle-specific fluorescent dyes. The images are divided into ten classes of organelles: DNA (Nuclei), ER (Endoplasmic reticulum), Giantin, (cis/medial Golgi), GPP130 (cis Golgi), Lamp2 (Ly-sosomes), Mitochondria, Nucleolin (Nucleoli), Actin, TfR (Endosomes), and Tubulin.
3. RN (RNAi data set [40]): this is a data set of 200 fluorescence microscopy images of fly cells (D. melanogaster) divided into ten classes. Each class contains 1024 x1024 TIFF images of phenotypes produced from one of ten knock-down genes, the IDs of which form the class labels.
4. MA (C. Elegans Muscle Age data set [40]): this data set is for classifying the age of the nematode given twenty-five images of C. elegans muscles collected at four ages representing the classes.
5. TB (Terminal Bulb Aging data set [40]): this is the companion data set to MA and contains 970 images of C. elegans terminal bulbs collected at seven ages rep-resenting the classes.
6. LY (Lymphoma data set [40]): this data set contains 375 images of malignant lymphoma representative of three types: CLL (chronic lymphocytic leukemia), FL (follicular lymphoma), and MCL (mantle cell lymphoma).
7. LG (Liver Gender Caloric Restriction (CR) data set [40]): this data set contains 265 images of liver tissue sections from six-month male and female mice on a CR diet; the two classes represent the gender of the mice.
8. LA (Liver Aging Ad-libitum data set [40]): this data set contains 529 images of liver tissue sections from female mice on an ad-libitum diet divided into four classes representing the age of the mice.
9. CO (Colorectal Cancer [41]): this is a Zenodo data set (record: 53169#.WaXjW8hJaUm) of 5000 histological images (150 x 150 pixels each) of human colorectal cancer divided into eight classes.
10. BGR (Breast Grading Carcinoma [42]): this is a Zenodo data set (record: 834910#.Wp1bQ-jOWUl) that contains 300 annotated histological images of twenty-one patients with invasive ductal carcinoma of the breast representing three classes/grades 1-3.
11. LAR (Laryngeal data set [43]): this is a Zenodo data set (record: 1003200#.WdeQcnBx0nQ) containing 1320 images of thirty-three healthy and early-stage cancerous laryngeal tissues representative of four tissue classes.
12. HP (set of immunohistochemistry images from the human protein atlas [44]): this is a Zenodo data set (record: 3875786#.XthkoDozY2w) of 353 images of fourteen proteins in nine normal reproductive tissues belonging to seven subcellular locations. The data set in [44] is partitioned into two folds, one for training (177 images) and one for testing (176 images).
13. RT (2D 3T3 Randomly CD-Tagged Images: Set 3 [45]): this collection of 304 2D 3T3 Randomly CD-Tagged images were created by randomly generating CD-tagged cell clones and imaging them by automated microscopy. The images are divided into ten classes. As in [45], the proteins are put into ten folds so that images in the training and testing sets never come from the same protein.
14. LO (Locate Endogenous data set [46]): this fairly balanced data set contains 502 images of endogenous cells divided into ten classes: Actin-cytoskeleton, Endosomes, ER, Golgi, Lysosomes, Microtubule, Mitochondria, Nucleus, Peroxisomes, and PM.
    This data set is archived at https://integbio.jp/dbcatalog/en/record/nbdc00296.

15. TR (Locate Transfected data set [46]): this is a companion data set to LO. TR contains 553 images divided into the same ten classes as LO but with the additional class of Cytoplasm for a total of eleven classes.

Data sets 1-8 can be downloaded at https://ome.grc.nia.nih.gov/iicbu2008/, data sets 9-12 can be found on Zenodo at https://zenodo.org/record/ by concatenating the data set's Zenodo record number provided in the descriptions above to this URL. Data set 13 is available at http://murphylab.web.cmu.edu/data/#RandTagAL, and data sets 14 and 15 are available on request. Unless otherwise noted, the five-fold cross-validation protocol is applied, and the Wilcoxon signed-rank test [47] is the measure used to validate experiments.

*4.2. Experimental results*

Reported in Tables 2-4 is the performance of the different activation functions on the CNN topologies Vgg16 and ResNet50, each trained with a batch size (BS) of 30 and a learning rate (LR) of 0.0001 for 20 epochs (the last fully connected layer has an LR 20 times larger than the rest of the layers (i.e., 0.002)), except the stochastic architectures that are trained for 30 epochs (because of slower convergence). The reason for selecting these settings was to reduce computation time. Images were augmented with random reflections on both axes and two independent random rescales of both axes by two factors uniformly sampled in [1,2] (using MATLAB data augmentation procedures). The objective was to rescale both the vertical and horizontal proportions of the new image. For each stochastic approach, a set of 15 networks was built and combined by sum rule. We trained the models using MATLAB 2019b; however, the pre-trained architectures of newer versions perform better.

The performance of the following ensembles is reported in Tables 4 and 5:
- ENS: sum rule of {MeLU ($k = 8$), Leaky ReLu, ELU, MeLU ($k = 4$), PReLU, SReLU, APLU, ReLu} (if $maxInput = 1$) or {MeLU ($k = 8$), MeLU ($k = 4$), SReLU, APLU, ReLu} (if $maxInput = 255$);
- eENS: sum rule of the methods that belong to ENS considering both $maxInput = 1$ and $maxInput = 255$;
- ENS_G: as in ENS but Small GaLU and GaLU are added, and in both cases $maxInput = 1$ or $maxInput = 255$;
- eENS_G: sum rule of the methods that belong to ENS_G but considering $maxInput = 1$ and $maxInput = 255$;
- ALL: sum rule among all the methods reported in table 4 with $maxInput = 1$ or $maxInput = 255$. Notice that when the methods with $maxInput = 255$ are combined, standard ReLu is also added to the fusion. Due to computation time, some activation functions are not combined with VGG16 and so are not considered;
- eALL: sum rule among all the methods, both with $maxInput = 1$ and $maxInput = 255$. Due to computation time, some activation functions are not combined with VGG16 and thus are not considered in an ensemble;
- 15ReLu: ensemble obtained by the fusion of 15 ReLU models. Each network is different because of the stochasticity of the training process;
- Selection: ensemble selected using SFFS (see section 3.1);
- Stoc_1: MeLU($k = 8$), leakyReLU, ELU, MeLU($k = 4$), PReLU, SReLU, APLU, GaLU, sGaLU. A $maxInput = 255$ has been used in the stochastic approach (see section 3.2);
- Stoc_2: the same nine functions of Stoc_1 and an additional set of seven activation functions: ReLU, SoftLearnable, PDeLU, learnableMish, SRS, SwishLearnable, and Swish. A $maxInput = 255$ has been used;
- Stoc_3: same as Stoc_2 but excluding all the activation functions proposed in [25-27] (i.e., Melu, GaLU, and sGaLU);
- Stoc_4: the ensemble detailed in section 3.

**Table 4**. Performance of activation function obtained using ResNet50.

|  | Activation | CH | HE | LO | TR | RN | TB | LY | MA | LG | LA | CO | BG | LAR | RT | HP | Avg |
|---|---|---|---|---|---|---|---|---|---|---|---|---|---|---|---|---|---|
| Resnet50 MaxInput=1 | MeLU (k=8) | 92.92 | 86.40 | 91.80 | 82.91 | 25.50 | 56.29 | 67.47 | 76.25 | 91.00 | 82.48 | 94.82 | 89.67 | 88.79 | 68.36 | 48.86 | 76.23 |
|  | Leaky ReLU | 89.23 | 87.09 | 92.80 | 84.18 | 34.00 | 57.11 | 70.93 | 79.17 | 93.67 | 82.48 | 95.66 | 90.33 | 87.27 | 69.72 | 45.45 | 77.27 |
|  | ELU | 90.15 | 86.74 | 94.00 | 85.82 | 48.00 | 60.82 | 65.33 | 85.00 | 96.00 | 90.10 | 95.14 | 89.33 | 89.92 | 73.50 | 40.91 | 79.38 |
|  | MeLU (k=4) | 91.08 | 85.35 | 92.80 | 84.91 | 27.50 | 55.36 | 68.53 | 77.08 | 90.00 | 79.43 | 95.34 | 89.33 | 87.20 | 72.24 | 51.14 | 76.48 |
|  | PReLU | 92.00 | 85.35 | 91.40 | 81.64 | 33.50 | 57.11 | 68.80 | 76.25 | 88.33 | 82.10 | 95.68 | 88.67 | 89.55 | 71.20 | 44.89 | 76.43 |
|  | SReLU | 91.38 | 85.58 | 92.60 | 83.27 | 30.00 | 55.88 | 69.33 | 75.00 | 88.00 | 82.10 | 95.66 | 89.00 | 89.47 | 69.98 | 42.61 | 75.99 |
|  | APLU | 92.31 | 87.09 | 93.20 | 80.91 | 25.00 | 54.12 | 67.20 | 76.67 | 93.00 | 82.67 | 95.46 | 90.33 | 88.86 | 71.65 | 48.30 | 76.45 |
|  | ReLU | 93.54 | 89.88 | 95.60 | 90.00 | 55.00 | 58.45 | **77.87** | 90.00 | 93.00 | 85.14 | 94.92 | 88.67 | 87.05 | 69.77 | 48.86 | 81.18 |
|  | Small GaLU | 92.31 | 87.91 | 93.20 | 91.09 | 52.00 | 60.00 | 72.53 | 90.00 | 95.33 | 87.43 | 95.38 | 87.67 | 88.79 | 67.57 | 44.32 | 80.36 |
|  | GaLU | 92.92 | 88.37 | 92.20 | 90.36 | 41.50 | 57.84 | 73.60 | 89.17 | 92.67 | 88.76 | 94.90 | 90.33 | 90.00 | 72.98 | 48.86 | 80.29 |
|  | Flexible MeLU | 91.69 | 88.49 | 93.00 | 91.64 | 38.50 | 60.31 | 73.33 | 88.33 | 95.67 | 87.62 | 94.72 | 89.67 | 86.67 | 67.35 | 44.32 | 79.42 |
|  | TanELU | 93.54 | 86.16 | 90.60 | 90.91 | 40.00 | 58.56 | 69.60 | 86.25 | 95.33 | 83.05 | 94.80 | 87.67 | 86.89 | 73.95 | 43.18 | 78.69 |
|  | 2D MeLU | 91.69 | 87.67 | 93.00 | 91.64 | 48.00 | 60.41 | 72.00 | 91.67 | 96.00 | 88.38 | 95.42 | 89.00 | 87.58 | 70.53 | 42.61 | 80.37 |
|  | MeLU+GaLU | 93.23 | 88.02 | 93.40 | 92.91 | 54.50 | 59.18 | 72.53 | 89.58 | 95.33 | 86.29 | 95.34 | 88.64 | 88.64 | 69.29 | 43.18 | 80.67 |
|  | splash | 93.54 | 87.56 | 93.80 | 90.00 | 47.50 | 55.98 | 72.00 | 82.92 | 94.33 | 84.19 | 95.02 | 86.00 | 87.12 | **75.70** | 42.61 | 79.21 |
|  | Symmetric GaLU | 93.85 | 84.19 | 92.80 | 89.45 | 47.50 | 58.66 | 72.80 | 87.08 | 95.33 | 82.67 | 94.44 | 87.33 | 87.80 | 71.52 | 52.84 | 79.88 |
|  | Symmetric MeLU | 92.62 | 86.63 | 92.40 | 89.27 | 50.00 | 60.62 | 72.27 | 85.42 | 95.00 | 85.14 | 94.72 | 90.00 | 87.58 | 66.71 | 50.57 | 79.93 |
|  | Soft Learnable v2 | 93.93 | 87.33 | 93.60 | 92.55 | 46.00 | 60.31 | 69.07 | 89.58 | 94.67 | 86.10 | 95.00 | 89.67 | 87.05 | 73.72 | 54.55 | 80.87 |
|  | Soft Learnable | 94.15 | 87.44 | 93.40 | 90.36 | 47.00 | 59.18 | 67.73 | 88.33 | 95.00 | 85.52 | 95.52 | 89.33 | 88.26 | 72.04 | 46.59 | 79.99 |
|  | PDELU | 94.15 | 87.21 | 92.00 | 91.64 | 51.50 | 56.70 | 70.93 | 89.58 | 96.33 | 86.67 | 95.08 | 89.67 | 88.18 | 72.76 | 46.59 | 80.59 |
|  | Mish | 95.08 | 87.56 | 93.20 | 91.82 | 45.00 | 58.45 | 69.07 | 86.67 | 95.33 | 86.67 | 95.48 | 90.00 | 88.41 | 53.41 | 34.09 | 78.01 |
|  | SRS | 93.23 | 88.84 | 93.40 | 91.09 | 51.50 | 60.10 | 69.87 | 88.75 | 95.00 | 86.48 | 95.72 | 88.33 | 89.47 | 54.06 | 48.86 | 79.64 |
|  | Swish Learnable | 93.54 | 87.91 | 94.40 | 91.64 | 48.00 | 59.28 | 69.33 | 88.75 | 95.33 | 83.24 | 96.10 | 90.00 | 89.32 | 41.15 | 39.77 | 77.85 |
|  | Swish | 94.15 | 88.02 | 94.20 | 90.73 | 48.50 | 59.90 | 70.13 | 89.17 | 92.67 | 86.10 | 95.66 | 87.67 | 87.65 | 65.05 | 32.39 | 78.79 |
|  | ENS | 95.38 | 89.53 | 97.00 | 89.82 | **59.00** | 62.78 | 76.53 | 86.67 | 96.00 | **91.43** | 96.60 | **91.00** | 89.92 | 74.00 | 50.00 | 83.04 |
|  | ENS_G | 93.54 | 90.70 | **97.20** | 92.73 | 56.00 | 63.92 | 77.60 | 90.83 | 96.33 | **91.43** | 96.42 | 90.00 | 90.00 | 73.76 | 50.00 | 83.36 |
|  | ALL | **97.23** | **91.16** | **97.20** | **95.27** | 58.00 | **65.15** | 76.80 | **92.92** | **98.00** | 90.10 | **96.58** | 90.00 | **90.38** | 74.67 | **53.98** | **84.49** |
| Resnet50 MaxInput=255 | MeLU (k=8) | 94.46 | 89.30 | 94.20 | 92.18 | 54.00 | 61.86 | 75.73 | 89.17 | 97.00 | 88.57 | 95.60 | 87.67 | 88.71 | 72.09 | 52.27 | 82.18 |
|  | MeLU (k=4) | 92.92 | 90.23 | 95.00 | 91.82 | 57.00 | 59.79 | 78.40 | 87.50 | 97.33 | 85.14 | 95.72 | 89.33 | 88.26 | 66.20 | 48.30 | 81.52 |
|  | SReLU | 92.31 | 89.42 | 93.00 | 90.73 | 56.50 | 59.69 | 73.33 | 91.67 | **98.33** | 88.95 | 95.52 | 89.67 | 87.88 | 68.94 | 48.30 | 81.61 |
|  | APLU | 95.08 | 89.19 | 93.60 | 90.73 | 47.50 | 56.91 | 75.20 | 89.17 | 97.33 | 87.05 | 95.68 | 89.67 | 89.47 | 71.44 | 51.14 | 81.27 |
|  | Small GaLU | 93.54 | 87.79 | 95.60 | 89.82 | 55.00 | 63.09 | 76.00 | 90.42 | 95.00 | 85.33 | 95.08 | 89.67 | 89.77 | 72.14 | 45.45 | 81.58 |
|  | GaLU | 92.92 | 87.21 | 92.00 | 91.27 | 47.50 | 60.10 | 74.13 | 87.92 | 96.00 | 86.86 | 95.56 | 89.33 | 87.73 | 70.26 | 44.32 | 80.20 |
|  | Flexible MeLU | 92.62 | 87.09 | 91.60 | 91.09 | 48.50 | 57.01 | 69.60 | 86.67 | 95.00 | 87.81 | 95.26 | 89.00 | 88.11 | 70.83 | 46.59 | 79.78 |
|  | 2D MeLU | 95.08 | 90.23 | 93.00 | 91.45 | 54.00 | 57.42 | 69.60 | 90.42 | 96.00 | 87.43 | 91.84 | 87.67 | 90.76 | 73.44 | 54.55 | 81.52 |
|  | MeLU+GaLU | 93.23 | 87.33 | 92.20 | 90.91 | 54.00 | 58.66 | 73.87 | 89.58 | 95.33 | 88.76 | 95.42 | 86.33 | 86.74 | 70.91 | 48.86 | 80.92 |
|  | splash | **96.00** | 87.67 | 92.80 | 93.82 | 50.50 | 60.62 | 78.13 | 89.58 | 96.67 | 87.81 | 95.18 | 90.33 | 91.36 | 68.81 | 51.70 | 82.06 |
|  | Symmetric GaLU | 92.00 | 85.58 | 91.20 | 89.64 | 43.50 | 57.94 | 70.93 | 79.58 | 91.33 | 85.14 | 95.34 | 87.33 | 85.98 | 69.37 | 47.16 | 78.13 |
|  | Symmetric MeLU | 92.92 | 88.37 | 93.40 | 92.00 | 44.00 | 58.56 | 69.60 | 91.67 | 93.33 | 84.00 | 94.94 | 87.33 | 88.79 | 70.30 | 44.89 | 79.60 |
|  | ENS | 93.85 | **91.28** | 96.20 | 93.27 | 59.00 | 63.30 | 77.60 | 91.67 | 98.00 | 87.43 | 96.30 | 89.00 | 89.17 | 71.11 | 50.00 | 83.14 |
|  | ENS_G | 95.08 | **91.28** | 96.20 | 94.18 | **63.00** | 64.85 | 78.67 | 92.50 | 97.67 | 87.62 | 96.54 | **89.67** | 89.77 | 71.36 | 51.14 | 83.96 |

| | | CH | HE | LO | TR | RN | TB | LY | MA | LG | LA | CO | BG | LAR | RT | HP | Avg |
|---|---|---|---|---|---|---|---|---|---|---|---|---|---|---|---|---|---|
| | ALL | **96.00** | 91.16 | **96.60** | **94.55** | 60.50 | 64.74 | 77.60 | **92.92** | 97.67 | **89.52** | **96.62** | 89.33 | **90.68** | 74.37 | 52.27 | 84.30 |
| eENS | | 94.77 | **91.40** | 97.00 | 92.91 | 60.00 | 64.74 | 77.87 | 88.75 | **98.00** | 90.10 | 96.50 | **90.00** | 89.77 | 73.23 | 50.57 | 83.70 |
| eENS_G | | 95.08 | 91.28 | 96.80 | 93.45 | **62.50** | **65.26** | 78.93 | 91.67 | 96.67 | **90.48** | 96.60 | 89.33 | 89.85 | 73.60 | 50.00 | 84.10 |
| eALL | | 96.92 | 91.28 | 97.20 | **95.45** | 60.50 | 64.64 | 77.87 | **93.75** | 97.67 | 90.10 | 96.58 | 89.67 | **90.68** | **74.37** | 52.27 | **84.59** |
| 15ReLu | | 95.40 | 91.10 | 96.20 | 95.01 | 58.50 | 64.80 | 76.00 | 92.90 | 97.30 | 89.30 | 96.30 | 90.00 | 90.04 | 73.00 | 50.57 | 83.76 |
| Selection | | 96.62 | 91.40 | 97.00 | 95.09 | 60.00 | 64.85 | 77.87 | **93.75** | 98.00 | 90.29 | **96.78** | 90.00 | 90.98 | 74.04 | 54.55 | 84.74 |
| Stoc_1 | | 97.81 | 91.51 | 96.66 | 95.87 | 60.04 | 65.83 | 80.02 | 92.96 | 99.09 | 91.24 | 96.61 | 90.77 | 91.03 | 74.20 | 50.57 | 84.95 |
| Stoc_2 | | 98.82 | 93.42 | 97.87 | 96.48 | **65.58** | **66.92** | **85.65** | 92.94 | 99.77 | **94.33** | 96.63 | 91.36 | 92.34 | **76.83** | 54.55 | **86.89** |
| Stoc_3 | | **99.43** | **93.93** | **98.04** | 96.06 | 64.55 | 66.41 | 83.24 | 90.04 | 96.04 | 93.93 | 96.72 | **92.05** | 91.34 | 75.89 | 51.70 | 85.95 |
| Stoc_4 | | 98.77 | 92.09 | 97.40 | 96.55 | 63.00 | 67.01 | 81.87 | 93.33 | **100** | 93.52 | 96.72 | 93.00 | 92.27 | 76.38 | 51.70 | 86.24 |

The most relevant results reported in Table 4 on ResNet50 can be summarized as follows:

- Ensemble methods outperform stand-alone networks. This result confirms previous research showing that changing activation functions is a viable method for creating ensembles of networks. Note how well 15ReLu outperforms (p-value of 0.01) the stand-alone ReLu;
- Among the stand-alone ResNet50 networks, ReLU is not the best activation function. The two activations that reach the highest performance on ResNet50 are MeLU ($k = 8$) with $maxInput = 255$ and Splash with $maxInput = 255$. According to the Wilcoxon Signed Rank Test, MeLU ($k = 8$) with $maxInput = 255$ outperforms ReLU with a p-value of 0.1. There is no statistical difference between MeLU ($k = 8$) and Splash (with $maxInput = 255$ for both);
- According to the Wilcoxon Signed Rank Test, Stoc_4 and Stoc_2 are similar in performance, and both outperform the other stochastic approach with a p-value of 0.1;
- Stoc_4 outperforms eALL, 15ReLu, and Selection with a p-value of 0.1. Selection outperforms 15ReLu with p-value of 0.01, but Selection's performance is similar to eALL.

**Table 5**. Activation performance on Vgg16.

| | Activation | CH | HE | LO | TR | RN | TB | LY | MA | LG | LA | CO | BG | LAR | RT | HP | Avg |
|---|---|---|---|---|---|---|---|---|---|---|---|---|---|---|---|---|---|
| Vgg16 | MeLU (k=8) | **99.69** | 92.09 | 98.00 | 92.91 | 59.00 | 60.93 | 78.67 | 87.92 | **86.67** | 93.14 | 95.20 | 89.67 | 90.53 | 73.73 | 42.61 | 82.71 |
| MaxInput=1 | Leaky ReLU | 99.08 | 91.98 | 98.00 | 93.45 | 66.50 | 61.13 | 80.00 | 92.08 | **86.67** | 91.81 | 95.62 | 91.33 | 88.94 | 74.86 | 38.07 | 83.30 |
| | ELU | 98.77 | 93.95 | 97.00 | 92.36 | 56.00 | 59.69 | 81.60 | 90.83 | 78.33 | 85.90 | 95.78 | 93.00 | 90.45 | 71.55 | 40.91 | 81.74 |
| | MeLU (k=4) | 99.38 | 91.16 | 97.60 | 92.73 | 64.50 | 62.37 | 81.07 | 89.58 | 86.00 | 89.71 | 95.82 | 89.67 | **93.18** | 75.20 | 42.61 | 83.37 |
| | PReLU | 99.08 | 90.47 | 97.80 | 94.55 | 64.00 | 60.00 | 81.33 | 92.92 | 78.33 | 91.05 | 95.80 | 92.67 | 90.38 | 73.74 | 35.23 | 82.49 |
| | SReLU | 99.08 | 91.16 | 97.00 | 93.64 | 65.50 | 60.62 | 82.67 | 90.00 | 79.33 | 93.33 | 96.10 | 94.00 | 92.58 | **76.80** | 45.45 | 83.81 |
| | APLU | 99.08 | 92.33 | 97.60 | 91.82 | 63.50 | 62.27 | 77.33 | 90.00 | 82.00 | 92.38 | 96.00 | 91.33 | 90.98 | 76.58 | 34.66 | 82.52 |
| | ReLU | **99.69** | 93.60 | 98.20 | 93.27 | 69.50 | 61.44 | 80.80 | 85.00 | 85.33 | 88.57 | 95.50 | 93.00 | 91.44 | 73.68 | 40.34 | 83.29 |
| | Small GaLU | 98.46 | 91.63 | 97.80 | 91.35 | 64.50 | 59.79 | 80.53 | 89.58 | 77.33 | 92.76 | 95.70 | 91.67 | 91.97 | 72.63 | 44.32 | 82.66 |
| | GaLU | 98.46 | 94.07 | 97.40 | 92.36 | 65.00 | 59.07 | 81.07 | 92.08 | 75.67 | 93.71 | 95.68 | 88.67 | 91.74 | 75.81 | 39.20 | 82.66 |
| | Flexible MeLU | 97.54 | 94.19 | 96.60 | 94.91 | 59.00 | 62.68 | 77.07 | 90.00 | 89.00 | 91.81 | 95.94 | 92.67 | 89.92 | 72.15 | 38.64 | 82.80 |
| | TanELU | 97.85 | 93.14 | 97.00 | 92.36 | 61.00 | 61.44 | 72.80 | 89.17 | 77.33 | 91.62 | 95.28 | 89.67 | 90.23 | 72.84 | 43.75 | 81.69 |
| | 2D MeLU | 97.85 | 93.72 | 97.20 | 92.73 | 61.00 | 61.34 | 81.60 | 91.25 | **92.33** | 94.48 | 95.86 | 89.67 | 92.35 | 71.91 | 38.64 | 83.46 |
| | MeLU+GaLU | 98.15 | 93.72 | 98.20 | 93.64 | 60.00 | 60.82 | 77.60 | 92.08 | 81.00 | 93.14 | 95.54 | 92.33 | 89.47 | 75.60 | 47.16 | 83.23 |
| | splash | 97.85 | 92.79 | 97.80 | 92.18 | 58.50 | 62.06 | 75.73 | 88.33 | 83.67 | 85.90 | 95.02 | 91.67 | 90.15 | 74.29 | 42.05 | 81.86 |
| | Symmetric GaLU | 99.08 | 92.79 | 97.20 | 92.91 | 60.50 | 60.00 | 78.93 | 88.33 | 79.33 | 91.62 | 95.52 | 92.67 | 91.67 | 73.91 | 40.34 | 82.32 |
| | Symmetric MeLU | 98.46 | 92.91 | 96.60 | 92.18 | 56.50 | 59.69 | 74.93 | 90.00 | 85.00 | 87.05 | 94.76 | 90.33 | 90.68 | 72.87 | 41.48 | 81.56 |
| | Soft Learnable v2 | 95.69 | 87.91 | 94.60 | 93.45 | 34.50 | 55.57 | 50.67 | 77.50 | 64.67 | 29.71 | 94.08 | 67.67 | 92.35 | 68.96 | 35.80 | 69.54 |
| | Soft Learnable | 98.15 | 92.91 | 97.00 | 91.82 | 47.50 | 54.33 | 62.13 | 86.67 | 95.67 | 65.90 | 95.04 | 84.33 | 90.38 | 71.08 | 40.34 | 78.21 |
| | PDELU | 98.77 | 93.60 | 96.40 | 92.18 | 59.00 | 58.25 | 76.80 | 87.92 | 87.67 | 89.33 | 95.36 | 90.33 | 91.74 | 75.24 | 42.05 | 82.30 |
| | Mish | 96.31 | 90.70 | 94.60 | 93.64 | 18.50 | 46.80 | 54.13 | 66.67 | 73.67 | 56.38 | 93.88 | 80.00 | 82.73 | 73.89 | 44.32 | 71.08 |
| | SRS | 71.08 | 59.19 | 45.00 | 51.64 | 29.50 | 31.44 | 57.60 | 61.25 | 61.00 | 45.33 | 86.88 | 57.00 | 67.50 | 39.74 | 19.32 | 52.23 |
| | Swish Learnable | 97.54 | 91.86 | 97.00 | 93.64 | 43.50 | 54.64 | 66.67 | 87.08 | 81.00 | 79.43 | 94.46 | 81.00 | 85.23 | 70.02 | 35.23 | 77.22 |
| | Swish | 98.77 | 92.56 | 96.80 | 93.64 | 63.50 | 58.97 | 80.80 | 90.00 | 89.00 | 93.14 | 94.68 | 93.33 | 91.74 | 75.24 | 39.77 | 83.46 |
| | ENS | 99.38 | 93.84 | 98.40 | 95.64 | 68.00 | 65.67 | 85.07 | 92.08 | 85.00 | 96.38 | 96.74 | 94.33 | 92.65 | 75.55 | 44.89 | 85.57 |
| | ENS_G | 99.69 | 94.65 | 99.00 | 95.45 | 72.00 | 64.95 | 86.93 | 92.50 | 83.33 | 97.14 | 96.72 | 94.67 | 92.65 | 75.56 | 45.45 | 86.07 |
| | ALL | 99.69 | 95.35 | 98.80 | 95.45 | 72.00 | 66.80 | 84.00 | 94.17 | 85.67 | 97.14 | 96.66 | 95.00 | 93.18 | 75.85 | 48.30 | 86.53 |
| Vgg16 MaxInput=255 | MeLU (k=8) | 99.69 | 92.09 | 97.40 | 93.09 | 59.50 | 60.82 | 80.53 | 88.75 | 80.33 | 88.57 | 95.94 | 90.33 | 88.33 | 73.01 | 47.73 | 82.40 |
| | MeLU (k=4) | 99.38 | 91.98 | 98.60 | 92.55 | 66.50 | 59.59 | 84.53 | 91.67 | 88.00 | 94.86 | 95.46 | 93.00 | 93.03 | 72.21 | 38.64 | 84.00 |
| | SReLU | 98.77 | 93.14 | 97.00 | 92.18 | 65.00 | 62.47 | 77.60 | 89.58 | 76.00 | 96.00 | 95.84 | 94.33 | 89.85 | 74.04 | 42.61 | 82.96 |
| | APLU | 98.77 | 92.91 | 97.40 | 93.09 | 63.00 | 57.32 | 82.67 | 90.42 | 77.00 | 90.67 | 94.90 | 93.00 | 91.21 | 75.65 | 36.36 | 82.29 |
| | Small GaLU | 99.38 | 92.91 | 97.00 | 92.73 | 50.50 | 62.16 | 78.40 | 90.42 | 73.00 | 94.48 | 95.32 | 92.00 | 90.98 | 73.61 | 42.61 | 81.70 |
| | GaLU | 98.77 | 92.91 | 97.60 | 93.09 | 66.50 | 59.48 | 83.47 | 90.83 | 95.00 | 85.52 | 95.96 | 91.67 | 93.41 | 75.45 | 38.64 | 83.88 |
| | Flexible MeLU | 99.08 | 95.00 | 97.20 | 93.45 | 62.00 | 55.98 | 76.80 | 89.17 | 83.00 | 88.57 | 95.64 | 91.33 | 91.29 | 73.00 | 37.50 | 81.93 |
| | MeLU+GaLU | 98.46 | 94.42 | 96.80 | 92.00 | 54.50 | 60.82 | 79.73 | 90.83 | 78.67 | 93.33 | 96.26 | 89.67 | 91.14 | 74.79 | 40.34 | 82.11 |
| | Symmetric GaLU | 97.85 | 92.21 | 97.40 | 93.64 | 58.00 | 58.14 | 73.87 | 91.67 | 79.33 | 91.43 | 95.18 | 90.33 | 89.55 | 74.47 | 34.09 | 81.14 |
| | Symmetric MeLU | 98.46 | 92.33 | 96.80 | 92.18 | 56.50 | 61.24 | 75.47 | 89.17 | 82.00 | 88.00 | 95.32 | 92.67 | 88.86 | 74.27 | 38.07 | 81.42 |
| | ENS | 99.38 | 93.84 | 98.80 | 95.27 | 68.50 | 64.23 | 84.53 | 92.50 | 81.33 | 96.57 | 96.66 | 95.00 | 92.20 | 75.27 | 43.75 | 85.18 |
| | ENS_G | 99.38 | 94.88 | 98.80 | 95.64 | 70.50 | 65.88 | 85.87 | 93.75 | 81.67 | 96.38 | 96.70 | 95.67 | 92.80 | 75.26 | 44.32 | 85.83 |
| | ALL | 99.69 | 95.47 | 98.40 | 95.45 | 70.00 | 63.92 | 83.73 | 94.17 | 82.67 | 96.38 | 96.60 | 95.00 | 92.73 | 75.78 | 45.45 | 85.69 |
| eENS | | 99.38 | 94.07 | 98.80 | 95.64 | 69.00 | 65.88 | 85.87 | 93.33 | 82.67 | 96.57 | 96.88 | 95.33 | 92.50 | 74.99 | 43.18 | 85.60 |
| eENS_G | | 99.69 | 94.65 | 99.00 | 95.27 | 70.50 | 65.57 | 86.93 | 92.92 | 83.33 | 97.71 | 96.82 | 95.00 | 92.42 | 76.09 | 44.32 | 86.01 |
| eALL | | 99.69 | 95.70 | 98.80 | 95.45 | 71.50 | 65.98 | 83.73 | 94.58 | 85.67 | 96.38 | 96.70 | 95.00 | 92.50 | 75.42 | 47.16 | 86.28 |
| 15ReLu | | 99.08 | 95.35 | 98.60 | 94.91 | 64.50 | 64.64 | 79.20 | 95.00 | 83.00 | 92.76 | 96.38 | 94.00 | 92.42 | 74.34 | 50.57 | 84.98 |
| Selection | | 99.69 | 95.26 | 98.60 | 94.91 | 71.00 | 64.85 | 86.67 | 94.58 | 84.67 | 95.24 | 96.72 | 94.33 | 93.56 | 75.48 | 47.16 | 86.18 |
| Stoc_4 | | 99.69 | 96.05 | 98.60 | 95.27 | 74.50 | 67.53 | 83.47 | 95.00 | 84.00 | 95.62 | 96.78 | 92.67 | 93.48 | 74.87 | 51.70 | 86.61 |

The most relevant results reported in Table 5 on Vgg16 can be summarized as follows:
- Again, the ensemble methods outperform the stand-alone CNNs. As was the case with ResNet50, 15ReLu strongly outperforms (p-value of 0.01) the stand-alone CNNs with ReLu;
- Among the stand-alone Vgg16 networks, ReLU is not the best activation function. The two activations that reach the highest performance on V6616 are MeLU ($k = 4$) with

$maxInput = 255$ and GaLU with $maxInput = 255$. According to the Wilcoxon Signed Rank Test, there is no statistical difference between ReLU, MeLU($k = 4$)-MI = 255, and GaLu-MI= 255;
- Interestingly, ALL with $maxInput = 1$ outperforms eALL with p-value of 0.05;
- Stoc_4 outperforms 15ReLu with p-value of 0.01, but the performance of Stoc_4 is similar to eALL, ALL ($maxInput = 1$), and Selection.

**5. Conclusions**

The goal of this work was to evaluate the performance of CNN ensembles by replacing the ReLU layers with activations from a large set of activation functions, including six new activation functions introduced here named 2D Mexican ReLU, TanELU, MeLU+GaLU, Symmetric MeLU, Symmetric GaLU and Flexible MeLU. Tests were run on two different networks: Vgg16 and ResNet50 across fifteen challenging image data sets representing various tasks. Different methods of making ensembles of the CNNs were also explored.

Experiments show that an ensemble of multiple CNNs that differ only in their activation functions outperforms the results of single CNNs. Experiments also show that among the single architectures there is no clear winner.

More studies need to investigate the performance gains of generating larger ensembles composed of different CNN architectures using many activation functions across even more data sets. Studies like the one presented here are difficult because investigating ensembles of CNNs requires enormous computational resources.

**Author Contributions:** L.N. conceived the presented idea. G.M., M.P. and L.N. performed the experiments. S.B., G.M. L.N. wrote the manuscript. M.P. and S.B. provided some resources.

**Funding:** This research received no external funding.

**Acknowledgments:** The authors are grateful to NVIDIA Corporation for supporting this research with the donation of a Titan Xp GPU. The authors also wish to acknowledge TCSC - Tampere Center for Scientific Computing and CSC - IT Center for Science (Finland) for generous computational resources.

**Conflicts of Interest:** The authors declare no conflicts of interest.